\documentclass[letterpaper]{article} 
\usepackage{aaai23}  
\usepackage{times}  
\usepackage{helvet}  
\usepackage{courier}  
\usepackage[hyphens]{url}  
\usepackage{graphicx} 
\usepackage{natbib}  
\usepackage{caption} 
\usepackage{times}  
\usepackage{helvet}  
\usepackage{courier}  
\usepackage[hyphens]{url}  
\usepackage{graphicx} 
\usepackage{natbib}  
\usepackage{caption} 
\usepackage[utf8]{inputenc} 
\usepackage[T1]{fontenc}    
\usepackage{url}            
\usepackage{booktabs}       
\usepackage{amsfonts}       
\usepackage{nicefrac}       
\usepackage{microtype}      
\usepackage{xcolor}         
\usepackage{multirow}
\usepackage{graphicx}
\usepackage{enumitem}
\usepackage{subfigure}
\usepackage{amssymb}
\usepackage{fdsymbol}
\usepackage{cancel}
\usepackage{xcolor}
\usepackage{wrapfig}
\usepackage{caption}
\usepackage{blindtext}
\usepackage{float}
\usepackage{stfloats}
\newtheorem{myDef}{Definition}

\newtheorem{myPro}{Problem}
\usepackage[ruled,linesnumbered,lined]{algorithm2e}
\usepackage{amsmath}
\usepackage{enumitem}
\usepackage{amssymb}
\usepackage{float}

\usepackage{newfloat}
\usepackage{listings}
\DeclareCaptionStyle{ruled}{labelfont=normalfont,labelsep=colon,strut=off} 
\floatstyle{ruled}
\newfloat{listing}{tb}{lst}{}
\floatname{listing}{Listing}
\title{Directed Acyclic Graph Structure Learning from Dynamic Graphs}
\author{
    Shaohua Fan\textsuperscript{1}, Shuyang Zhang\textsuperscript{1}, Xiao Wang\textsuperscript{1,2}, Chuan Shi\textsuperscript{1,2}\thanks{Corresponding author.}
}
\affiliations{
    \textsuperscript{1}Beijing University of Posts and Telecommunications, China\\
    \textsuperscript{2}Peng Cheng Laboratory, China

   \{fanshaohua, sonyazhang, xiaowang, shichuan\}@bupt.edu.cn
}

\usepackage{bibentry}

\begin{document}

\maketitle

\begin{abstract}
Estimating the structure of directed acyclic graphs (DAGs) of features (variables) plays a vital role in revealing the latent data generation process and providing causal insights in various applications. Although there have been many studies on structure learning with various types of data, the structure learning on the dynamic graph has not been explored yet, and thus we study the learning problem of node feature generation mechanism on such ubiquitous dynamic graph data. In a dynamic graph, we propose to simultaneously estimate contemporaneous relationships and time-lagged interaction relationships between the node features. These two kinds of relationships form a DAG, which could effectively characterize the feature generation process in a concise way. To learn such a DAG, we cast the learning problem as a continuous score-based optimization problem, which consists of a differentiable score function to measure the validity of the learned DAGs and a smooth acyclicity constraint to ensure the acyclicity of the learned DAGs. These two components are translated into an unconstraint augmented Lagrangian objective which could be minimized by mature continuous optimization techniques. The resulting algorithm, named GraphNOTEARS, outperforms baselines on simulated data across a wide range of settings that may encounter in real-world applications. We also apply the proposed approach on two dynamic graphs constructed from the real-world Yelp dataset, demonstrating our method could learn the connections between node features, which conforms with the domain knowledge.
\end{abstract}

\section{Introduction}
\par A Bayesian network (BN) is a probabilistic graphical model that represents a set of variables and their conditional dependencies. It has been widely used in machine learning applications~\cite{pearl2011bayesian,ott2003finding,friedman1997bayesian}. The structure of a BN takes the form of a directed acyclic graph (DAG) and provides a convenient and interpretable output which is needed in today’s high-stake applications of artificial intelligence, such as healthcare, finance and autonomous driving. The edges in a DAG represent the directed generation relationships between variables (e.g., features) in a system. When these edges are not known based on prior knowledge, one possible solution is to resort to DAG
structure learning, namely, learning the
edges in a graphical model from the observed data.
\begin{figure}[t]
	\centering
	\includegraphics[width=8.5cm]{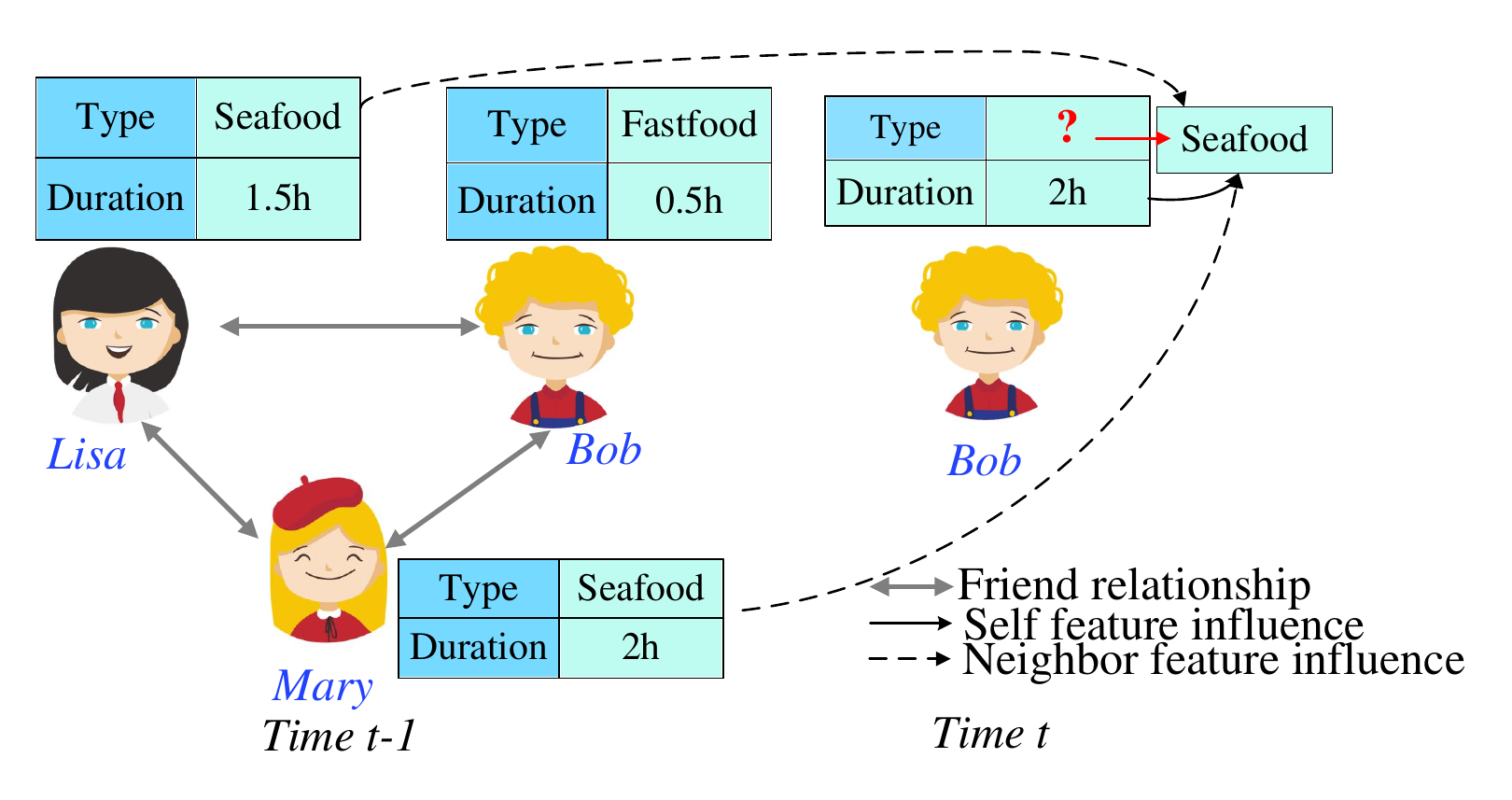}
	\caption{A toy example of feature generation process on a dynamic graph.}
	\label{fig:fig1}
\end{figure}

\par Existing approaches for DAG learning mostly focus on dealing with tabular data, i.e., each sample is independently drawn from the same distribution (IID data)~\cite{spirtes2000causation,neuberg2003causality,spirtes2013causal,geiger1994learning,heckerman1995learning}. Nevertheless, in real-world scenarios, there usually exists associations between samples, so the generation of features of certain samples may be influenced by the other samples through the links between samples. Several pioneer works have proposed constraint-based methods to learn DAGs from static (i.e., equilibrium) graph data~\cite{maier2010learning,lee2016learning,maier2013sound}. They usually test the conditional independencies of the attributes of entities in graph data to form DAGs among those variables. 

\par However, many real-world scenarios exhibit temporal information in graph data. For example, as shown in Fig.~\ref{fig:fig1}, Bob's friends, Lisa and Mary, went to eat seafood and posted recommendations and the duration of the meal on social media at timestamp $t-1$. Bob viewed this information. When he has enough available time at timestamp $t$, he will choose to eat seafood with high probability. Hence, the generation of the value of Bob's meal type at timestamp $t$ will be both determined by the current available time and recommendation from social network at the previous timestamp. Another example is that the risk of ones to be infected by COVID-19 may be both determined by current protection status (e.g., wearing a mask or keeping social distance) and the ratio of neighborhood who has been vaccinated or infected at previous timestamp two weeks ago. Actually, the current methods largely ignore modeling the temporal interaction, so the true data generation process could not be revealed accurately.

\par When learning DAGs from dynamic graphs, there are two intractable challenges we need to confront. First, a dynamic graph contains complex temporal interactions between samples, so what kind of DAGs would reflect the generation process of features in a dynamic graph?  As the samples at each timestamp will be generated based on the interactions from previous timestamps, the learned DAG should model the generation process of each new sample at each timestamp, considering the time-lagged interaction information. Second, how to efficiently learn a DAG from complex evolutional graph data? Compared with IID data, a dynamic graph contains both temporal and interaction information, hence it has a more complicated data generation mechanism. It is non-trivial to design a DAG learning method for dynamic graphs. Fortunately, owing to the well-developed optimization techniques, it is possible to develop a differentiable score function that could measure the validity of candidate DAGs and resort to blackbox solvers to find the optimal DAG efficiently.

\par Particularly, to address these two challenges, we propose an effective score-based approach for learning DAGs that could scale gracefully to dynamic graphs with high-dimensional node features, called GraphNOTEARS. To solve the first challenge, we propose to learn an intra-slice matrix to characterize contemporaneous relationships between variables, and several inter-slice matrices to characterize multi-step time-lagged graph influence on current timestamp. Meanwhile, an acyclicity constraint is required to ensure the acyclicity of the learned whole graph. As for the second challenge, we cast the problem as a score-based optimization problem, and develop a least-squares loss based score function. The score function leverages the temporal and 
interaction information, as well as two kinds of learnable structural matrices to reconstruct the data. With the smooth acyclicity constraint, we translate the original unsolvable constraint problem into an unconstraint augmented Lagrangian objective. The resulting program could be solved by the standard second-order optimization schemes efficiently. The main contributions of this paper are summarized as follows:
\begin{itemize}[leftmargin = 10 pt]
    \item To our best knowledge, we first study the DAG learning problem on dynamic graphs. Because of the ubiquity of dynamic graph data in real applications, learning DAGs on such data could reveal the underlying feature generation process, provide skeletons for possible Bayesian networks, and answer causal questions, like the effect of various interventions. These applications are very important for building explainable, robust, and generalized algorithms.
    \item We develop a score-based learning method for simultaneously estimating the structure and parameters of a sparse DAG from a dynamic graph. The resulting method can be used to learn the relationships between variables of arbitrary time-lagged order in a dynamic graph, without any implicit assumptions on the underlying DAG topologies.
    \item We conduct extensive simulation experiments with broad range settings which may encounter in real world, validating the effectiveness of our approach in revealing the feature generation mechanism of dynamic graphs. The experiments on real-world datasets well demonstrate the rationality of the relationships inferenced by GraphNOTEARS.
\end{itemize}


%



\section{Background and Related Works}
A DAG $G$ is faithful with respect to a joint distribution $\mathcal{P}$ of a set of variables if all and only the conditional independencies of variables true in $\mathcal{P}$ are entailed by $G$~\cite{pearl2014probabilistic}. The faithfulness assumption
enables one to recover $G$ from $\mathcal{P}$. Given samples $D$ from an unknown distribution corresponding
to a faithful but unknown DAG, structure learning refers to recovering the DAG from $D$.

\par  Existing methods for DAG learning can be classified into constraint-based methods and score-based methods. Most constraint-based DAG learning methods~\cite{spirtes2000causation, neuberg2003causality, spirtes2013causal} first use conditional independence tests to find graph
skeleton and then determine the orientations of the edges up to the Markov equivalence class,
which usually contains DAGs that can be structurally diverse and may still have many unoriented edges. Score-based methods~\cite{geiger1994learning, huang2018generalized,hyvarinen2013pairwise}, on the other hand, define
a score function to find the best DAG that fits the
given data. Unlike constraint-based methods that assume faithfulness and identify only the Markov equivalence
class, these methods are able to distinguish between different DAGs in the same equivalence class, owing to the additional assumptions on data distribution and/or functional classes. However, due to the acyclicity constraint and superexponential in the number
of nodes of DAGs
to search over, score-based methods are computationally expensive. The recent work, NOTEARS~\cite{zheng2018dags}, expresses the acyclicity of a DAG by a smooth equality constraint under the linear assumption, which makes it possible to
formulate structure learning as a smooth minimization
problem subject to this equality constraint. DYNOTEARS~\cite{pamfil2020dynotears} extend NOTEARS to learn the DAG of time-series data, which incorporates the temporal information into the score function. 

The mainstream of existing methods is designed for the tabular samples, which are independently identically drawn from the same distribution.  However, in many real-world settings, there may exist links between samples and the links may influence the features generation process of samples. Maier et al.~\cite{maier2010learning} first extend the well-known PC algorithm to the relational setting for
learning causal relationships from relational data, called RPC. Later, Maier et
al.~\cite{maier2013sound} demonstrate the lack of completeness of RPC and introduce a sound and complete algorithm, named RCD. All of these relational DAG learning algorithms are constraint-based methods and they could only handle static graphs, ignoring temporal information. In this paper, we propose an efficient score-based method that could model the temporal interaction information.

\section{DAG Structure Learning on Dynamic Graphs}

\begin{figure*}[t]
	\centering
	\includegraphics[width=15cm]{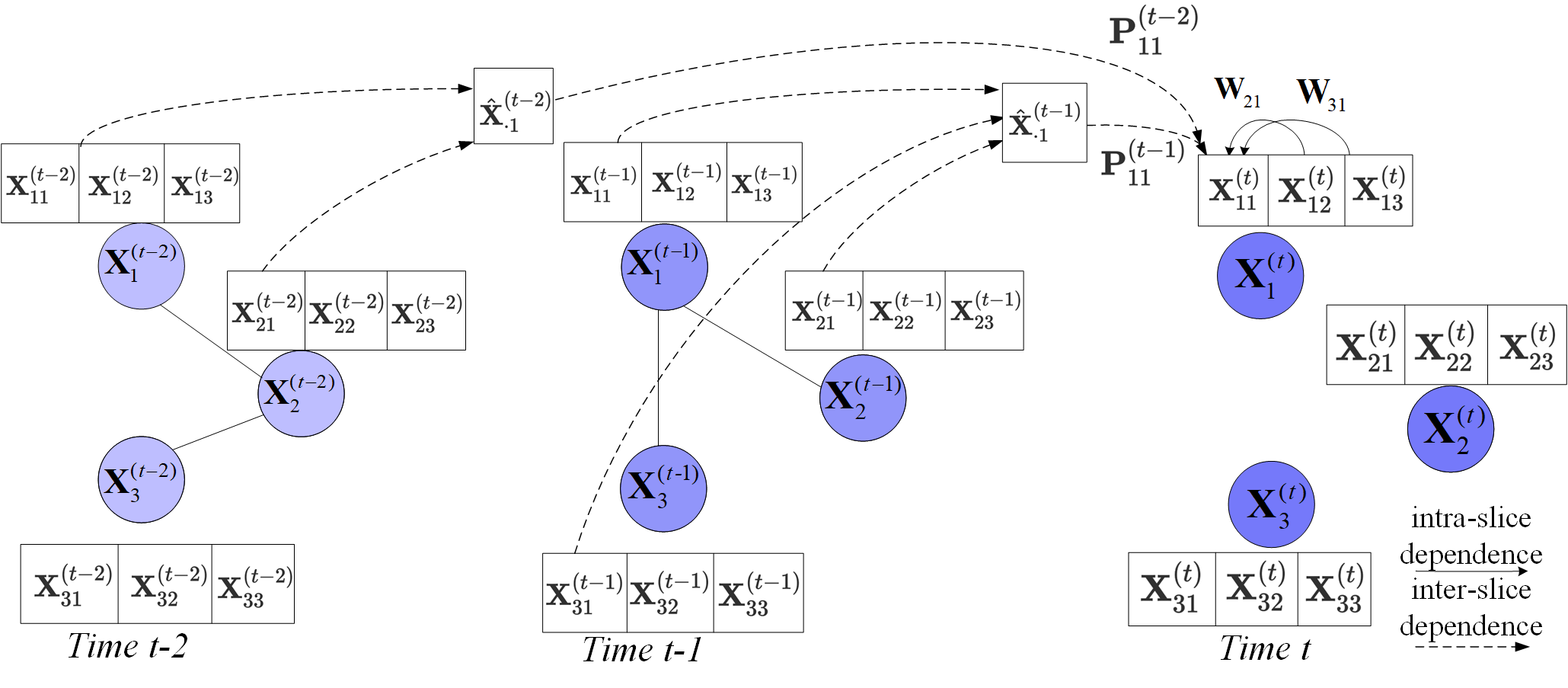}
	\caption{Illustration of intra-slice (solid lines) and inter-slice (dashed lines) dependencies in a dynamic graph with $n=3$ samples and $d = 3$ variables at each timestamp and time-lagged order $p = 2$. For clarity, we ignore the edges that do not influence the variables $\mathbf{X}_{11}^{(t)}$.}
	\label{fig:fig2}
	\vspace{-10pt}
\end{figure*}

In this section, we introduce the definition of dynamic graphs,
the problem of DAG structure learning on dynamic graphs as well as the proposed model: GraphNOTEARS.
\subsection{Problem Formulation}
\begin{myDef}[Dynamic Graph] A dynamic graph is $\mathcal{G}=\{(\mathbf{X}^{(1)}, \mathbf{A}^{(1)}), \cdots, (\mathbf{X}^{(T)}, \mathbf{A}^{(T)})\}$, where $T$ is the total number of timestamps, tuple ($\mathbf{X}^{(T)}$, $\mathbf{A}^{(T)}$) represents the graph at timestamp $T$, $\mathbf{X}^{(T)}\in\mathbb{R}^{n\times d}$ is the matrix of node features and  $\mathbf{A}^{(T)}\in\mathbb{R}^{n\times n}$ is the adjacency matrix of nodes, $n$ is the number of nodes\footnote{We assume the number of nodes at each timestamp remains the same.}, $d$ is the number of node features (i.e., variables). 
\end{myDef}
At each timestamp $t$, we assume that each feature of nodes is generated based on the contemporaneous variables and time-lagged neighborhood variables. For instance, as depicted in Fig.~\ref{fig:fig2}, the variable $\mathbf{X}_{11}^{(t)}$ of sample $\mathbf{X}_1^{(t)}$ at timestamp $t$ is determined by the contemporaneous variables $\mathbf{X}_{12}^{(t)}$ and $\mathbf{X}_{13}^{(t)}$ from the same sample\footnote{In this paper, we use node and sample interchangeably.} with coefficients $\mathbf{W}_{21}$ and $\mathbf{W}_{31}$, and the time-lagged aggregated neighborhood variables $\hat{\mathbf{X}}_{.1}^{(t-1)}$ and $\hat{\mathbf{X}}_{.1}^{(t-2)}$ from timestamp $t-1$ and $t-2$ with coefficients $\mathbf{P}^{(t-1)}_{11}$ and $\mathbf{P}^{(t-2)}_{11}$, respectively. We call these intra-slice and inter-slice dependencies, respectively. One may argue that the generation of variables of samples at current timestamp $t$ should also depend on the neighborhood samples at the same timestamp. In this paper, we assume that neighborhood behaviour needs a delay to influence ego nodes. The delay influence phenomenon is very common in real-world applications. One example is that the user’s preference on restaurant category may be influenced by their friends’ recommendation on Yelp platform, and the user will go to the restaurant in a few days. Moreover, we assume the interactions between samples are given at each timestamp, which is very common in dynamic graph data~\cite{rossi2020temporal,sankar2020dysat}. For example, we could easily get the following relationship between users on Yelp platform.

\par Moreover, we assume that in each timestamp, the effects of the time-lagged influences from neighborhood on current timestamp could be at most $p$ timestamps, where $p\le T$. We also make the stationary process assumption~\cite{hamilton2020time} that the generation process is fixed through time and is identical for generating each timestamp in the time-series, which is very common assumption in time-series data~\cite{hamilton2020time}. Without loss of any generality, we propose to model the data generation process at timestamp $t$ with the structural vector autoregressive (SVAR) model~\cite{demiralp2003searching,kilian2013structural}:

\begin{equation}
\begin{aligned}
  \mathbf{X}^{(t)} &= \mathbf{X}^{(t)}\mathbf{W} + \mathbf{\hat{A}}^{(t-1)}\mathbf{X}^{(t-1)}\mathbf{P}^{(t-1)}+\cdots\\
    &+\mathbf{\hat{A}}^{(t-p)}\mathbf{X}^{(t-p)}\mathbf{P}^{(t-p)}+\mathbf{Z},
\end{aligned}
    \label{Eq:Eq1}
\end{equation}
where $\mathbf{W}\in\mathbb{R}^{d\times d}$ and $\mathbf{P}^{(t-i)}(i\in\{1,\cdots, p\})\in\mathbb{R}^{d\times d}$ represents weighted adjacency matrices with nonzero entries corresponding to the intra-slice and inter-slice edges, respectively. $\mathbf{\mathbf{\hat{A}}}^{(t-i)}(i\in\{1,\cdots, p\})$ are normalized adjacency matrices, which is computed by $\mathbf{D}^{-\frac{1}{2}}(\mathbf{A}+\mathbf{I})\mathbf{D}^{-\frac{1}{2}}$, $\mathbf{D}_{ii}=\sum_j \mathbf{A}_{ij}$, $\mathbf{I}$ is the identity matrix of $\mathbf{A}$. The reason for $\mathbf{A}+\mathbf{I}$ is that the time-lagged influence should not only from neighborhood but also from itself at previous timestamps. And $\mathbf{Z}$ is a centered error matrix, where each row is independent for each sample. The error variables could be Gaussian, Exponential or others. The overall intuition of Eq. (\ref{Eq:Eq1}) would be that, the generation of the value of a new variable $i$ of $\mathbf{X}^{(t)}$ depends on two parts: parent variables from contemporaneous variables (i.e., $\mathbf{X}^{(t)}\mathbf{W}$) and parents variables from time-lagged neighborhood variables (i.e., $\mathbf{\hat{A}}^{(t-1)}\mathbf{X}^{(t-1)}\mathbf{P}^{(t-1)}+\cdots+\mathbf{\hat{A}}^{(t-p)}\mathbf{X}^{(t-p)}\mathbf{P}^{(t-p)}$). $\mathbf{X}^{(t)}\mathbf{W}$ represents the relationship between contemporaneous variables. According to SVAR model, contemporaneous variables usually exhibit a causal order, hence $\mathbf{W}$ represents the causal order of contemporaneous variables and is acyclic, where $\mathbf{W}_{kj}$ represents the coefficient of the parent variable $k$ at timestamp $t$ on the variable $j$ at the same timestamp. And $\mathbf{P}^{(t-i)}_{kj}$ represents the coefficient of $k$-th aggregated node variables at timestamp $t-i$ on $j$-th variable at timestamp $t$.  For the detailed pseudocode of Eq. (\ref{Eq:Eq1}), please refer to Appendix~A.2~\footnote{Supplementary material:https://drive.google.com/file/d/1S1pz\\EyyC9kNL6s97yQxRt5lOLdO5L8sz/view?usp=sharing}.

\par Let $\mathbf{M}=[\mathbf{X}^{(t-1)}| \cdots| \mathbf{X}^{(t-p)}]$ be the $n\times pd$ matrix of time-lagged node features data, $\mathbf{A}=[\mathbf{\hat{A}}^{(t-1)}| \cdots| \mathbf{\hat{A}}^{(t-p)}]$ be the $n\times pn$ matrix of time-lagged interaction data, and $\mathbf{P}=[\mathbf{P}^{(t-1)}| \cdots| \mathbf{P}^{(t-p)}]$ be the $pd \times d$ matrix of inter-slice weights. We could rewrite Eq. (\ref{Eq:Eq1}) as following compact form of structural equation model (SEM)~\cite{hox1998introduction}:

\begin{equation}
    \mathbf{X} = \mathbf{X}\mathbf{W} + \mathbf{A}\boxtimes\mathbf{M}\mathbf{P}+\mathbf{Z},
    \label{Eq:compact}
\end{equation}
where $\mathbf{A}\boxtimes\mathbf{M}=[\mathbf{\hat{A}}^{(t-1)}\mathbf{X}^{(t-1)}| \cdots| \mathbf{\hat{A}}^{(t-p)}\mathbf{X}^{(t-p)}]\in\mathbb{R}^{n\times pd}$. This general formulation covers scenarios in which the time-lagged data matrix $\mathbf{M}$ or $\mathbf{A}$ are
not a contiguous sequence of time slices (i.e., from $t - p$ to $t - 1$). For example, someone usually meets their friends in the weekend, hence one can include the lagged data matrix $\mathbf{M}$ or $\mathbf{A}$ only those time points that have an impact on the variables at timestamp $t$. Based on the generation formulation, we formulate our target problem:

\begin{myPro}[DAG Structure Learning on Dynamic Graph] Given a dynamic graph $\mathcal{G}$, the goal of DAG structure learning on the dynamic graph is to estimate the generation process of node features, which usually forms as a DAG, both considering the contemporaneous effect from each node itself and neighborhood effects from time-lagged graphs.
\end{myPro}

\par Hence, given the data $\mathbf{X}$, $\mathbf{M}$ and $\mathbf{A}$, the goal of this paper is to estimate weighted adjacency matrices $\mathbf{W}$ and $\mathbf{P}$, which could characterize the node feature generation process in a dynamic graph.
\subsection{The Proposed Model: GraphNOTEARS}
An SEM could be found through minimizing the least-squares (LS) loss~\cite{zheng2018dags}. The statistical properties of
the LS loss in scoring DAGs have been extensively studied: The minimizer of the least-squares loss provably recovers a
true DAG with high probability on finite-samples and in high-dimensions~\cite{aragam2015learning,loh2014high}. Inspired by this, we propose to estimate $\mathbf{W}$ and $\mathbf{P}$ by minimizing the following LS loss:
\begin{equation}
    \mathcal{L}(\mathbf{W}, \mathbf{P}) = \frac{1}{2n}||\mathbf{X}-\mathbf{X}\mathbf{W}-\mathbf{A}\boxtimes\mathbf{M}\mathbf{P}||^2_F.
\end{equation}
Moreover, the edges in $\mathbf{P}$ go only forward in time and
thus they do not create cycles. To ensure
that the whole graph is acyclic, it thus suffices to
require that $\mathbf{W}$ is acyclic. In this work, we utilize
acyclic constraint proposed by NOTEARS ~\cite{zheng2018dags} to ensure the acyclicity of learned  $\mathbf{W}$, which states that: a directed graph
$G$ with binary adjacency matrix $\mathbf{W}$ is acyclic if and only if:
\begin{equation}
    h(\mathbf{W}) := \text{trace}(e^{\mathbf{W}\circ \mathbf{W}})-d=0, 
\end{equation}
where $e^{\mathbf{W}\circ \mathbf{W}}$ is the matrix exponential of ${\mathbf{W}\circ \mathbf{W}}$, and $\circ$ denotes the Hadamard product of two matrices. To enforce the sparsity of $\mathbf{W}$ and $\mathbf{P}$, we also introduce $\ell_1$ penalties in the objective function. The overall optimization problem is:
\begin{equation}
   \begin{aligned}
    &\min_{\mathbf{W}, \mathbf{P}} f(\mathbf{W}, \mathbf{P})\\
     \text{s.t.}\quad & h(\mathbf{W}) := \text{trace}(e^{\mathbf{W}\circ \mathbf{W}})-d=0,\\
    \text{with}\quad & f(\mathbf{W}, \mathbf{P})=\frac{1}{2n}||\mathbf{X}-\mathbf{X}\mathbf{W}-\mathbf{A}\boxtimes\mathbf{M}\mathbf{P}||^2_F \\
    &+ \lambda_{\mathbf{W}}||\mathbf{W}||_1+ \lambda_{\mathbf{A}}||\mathbf{P}||_1,\\
    \end{aligned}
    \label{Eq:Final obj}
\end{equation}
where the $||\cdot||_1$ represents the element-wise $\ell_1$ norm, and $\lambda_{\mathbf{W}}$ and $\lambda_{\mathbf{A}}$ represent the coefficients of $||\mathbf{W}||_1$ and $||\mathbf{P}||_1$, respectively. As we can see, given $\mathbf{X}$, $\mathbf{M}$ and $\mathbf{A}$, though minimizing this objective, our model could simultaneously recover the contemporaneous dependencies $\mathbf{W}$ and neighborhood dependencies $\mathbf{P}$ from data, as well as ensure the acyclicity of the learned graph.
\subsection{Optimization}
The above optimization is a standard equality-constrained program (ECP). We translate
the problem to an unconstrained problem with the following smooth augmented Lagrangian objective
\begin{equation}
    \min_{\mathbf{W}, \mathbf{P}} f(\mathbf{W}, \mathbf{P}) + \frac{\rho}{2} h(\mathbf{W})^2+\alpha h(\mathbf{W}).
    \label{Eq:optimization}
\end{equation}

The resulting problem can be solved using efficient solvers such as L-BFGS-B~\cite{zhu1997algorithm}, and the update strategy of $\rho$ and $\alpha$ is the same as~\cite{zheng2018dags}. To reduce false discoveries~\cite{zhou2009thresholding}, we threshold the edge weights of $\mathbf{W}$ and $\mathbf{P}$ via hard thresholds $\tau_{\mathbf{W}}$ and $\tau_{\mathbf{P}}$: After obtaining a stationary point $\mathbf{W}$ and $\mathbf{P}$, given
a fixed threshold $\mathbf{\tau_{W}}>0$ and $\mathbf{\tau_{P}}>0$, set any weights smaller than $\mathbf{\tau_{W}}$ and $\mathbf{\tau_{P}}$ in absolute value to zero, termed as $\widetilde{\mathbf{W}}$ and $\widetilde{\mathbf{P}}$, and their binary version termed as $\mathbf{\hat{W}}$ and $\mathbf{\hat{P}}$, respectively.

\subsection{Discussion}
\label{Sec::Discuss}
Here we discuss the identifiability, some limitations and possible extensions of GraphNOTEARS.

\textbf{Identifiability} Identifiability is the key research problem of SVAR models of the econometrics literature~\cite{kilian2013structural}. Identifiability of structure learning on time-series data has been discussed in~\cite{pamfil2020dynotears} by using the conclusions of SVAR model. As a dynamic graph could be viewed as a special time-series data, where $\mathbf{A}\boxtimes\mathbf{M}$ in Eq. (\ref{Eq:compact}) is the aggregated time-lagged features, we assume the conditions for the identifiability in~\cite{pamfil2020dynotears} will be also held in our model, i.e., the error $\mathbf{Z}$ could be drawn from non-Gaussian or standard normal distribution. Thus, $\mathbf{W}$ and $\mathbf{P}$ of our model are identifiable under reasonable conditions.

\textbf{Assumptions} To be simplified, we have assumed that the structure of the process of variable generation is fixed across time and is identical for
all timestamps. Based on this assumption, when long time-series data is available, our model could easily utilize such long time-series in the objective function Eq. (\ref{Eq:Final obj}) by extending data matrices (i.e., $\mathbf{X}$, $\mathbf{M}$ and $\mathbf{A}$) to tensors and keeping the parameter matrices $\mathbf{W}$ and $\mathbf{P}$ unchanged. And all our experiments are based on this extension. This stationary process assumption is a very common assumption in time-series data~\cite{hamilton2020time} and could be relaxed in several ways. We could allow the directed dependency structure of data to vary smoothly over time~\cite{song2009time} or have discrete change points~\cite{grzegorczyk2011non}.
 
\textbf{Nonlinear relationship} As a very beginning work of structure learning on dynamic graphs, we follow previous works on structure learning~\cite{zheng2018dags,pamfil2020dynotears} that first consider the linear scenarios. Note that
linear assumption is made purely for simplicity,
so that our paper could focus on the most salient temporal and network aspects of this problem. Inspired by the GNNs~\cite{wu2020comprehensive,fan2019metapath,fan2020one2multi,fan2021generalizing,fan2022debiased,fan2022debiasing,wang2017community,chen2023universal} and the nonlinear structure learning methods~\cite{zheng2020learning, yu2019dag, lachapelle2019gradient}, we could model the nonlinear effects of neighbors by GNNs. 

An important feature of GraphNOTEARS is its simplicity, both in terms of formulating an objective function and optimizing it. Nevertheless, the proposed model is general enough to be extended to more complicated scenarios.
\section{Experiments}
\label{Sec::experiments}
It is notoriously hard to obtain the ground truth of causal structure because it is difficult to obtain the underlying data generation process of real-world problems. To validate the effectiveness of our method, in this section, we follow the setting in~\cite{zheng2018dags, pamfil2020dynotears, maier2010learning, maier2013sound}, which conduct extensive experiments on synthetic data with known generating mechanisms to simulate real-world scenarios. \footnote{Code and data: https://github.com/googlebaba/GraphNOTEARS.}

\textbf{Datasets.} To validate the effectiveness of GraphNOTEARS against existing approaches, we simulate data according to the SEM from Eq. (\ref{Eq:compact}). We need three steps to this process: 1) generating the weighted graphs $\mathcal{G}_\mathbf{W}$ and $\mathcal{G}_\mathbf{P}$, and adjacency matrix $\mathbf{A}$; 2) generating data matrices $\mathbf{X}$ and $\mathbf{M}$ based on $\mathcal{G}_\mathbf{W}$ and $\mathcal{G}_\mathbf{P}$; 3) running all algorithms on  all or partial of $\mathbf{X}$, $\mathbf{M}$ and $\mathbf{A}$ based on whether the model considers this kind of information and computing metrics respectively. Particularly, following~\cite{pamfil2020dynotears}, we use
either the Erd\H{o}s-R\'{e}nyi (ER) model~\cite{newman2018networks} or
the Barab\'{a}si-Albert (BA) model~\cite{barabasi1999emergence} to generate intra-slice graphs $\mathcal{G}_\mathbf{W}$. And for inter-slice graph $\mathcal{G}_\mathbf{P}$, we use ER model or Stochastic Block Model (SBM)~\cite{newman2018networks}. These graph generation models could simulate real-world variable generation processes, like ``rich get richer'', ``preferential attachment'' and ``cluster effect''.  To get weighted intra- and inter-slice matrices $\mathbf{W}$ and $\mathbf{P}$, we sample weights uniformly at random from $[-2, -0.5]\cup [0.5, 2]$. For the generation of $\mathbf{A}$, we connect each pair of samples with 0.1 probability. 
Once given $\mathbf{W}$, $\mathbf{P}$ and $\mathbf{A}$, we use the SEM from Eq. (\ref{Eq:compact}) to
generate a data matrix $\mathbf{X}$ of size $n\times d$. In particular, we first generate the variables of $\mathbf{X}$ and $\mathbf{M}$ based on the sorted topological order of $\mathbf{W}$ same as~\cite{zheng2018dags}. And we generate current timestamp observation according to: $\mathbf{X} = \mathbf{X}\mathbf{W} + \mathbf{A}\boxtimes\mathbf{M}\mathbf{P}+\mathbf{Z}$. For the noise term
$\mathbf{Z}$, we utilize Gaussian noise and Exponential noise. Moreover, to compare GraphNOTEARS against baselines with a wide range of sample sizes and the number of variables, we vary the sample size $n\in\{100, 200, 500\}$, the number of variables $d\in\{5,10,20,30\}$ at each timestamp, and the length of time-series $T$ is set as 7 for all experiments. A detailed introduction to the data generation process is in Appendix A.

\textbf{Baselines.} Because we study a new problem, there is no baseline specially designed for this problem. We compare the following two alternatives that could deal with the problem in an indirect way.

\begin{itemize}[leftmargin = 10 pt]
    \item NOTEARS~\cite{zheng2018dags}+LASSO: This is a two-step approach. We use static
NOTEARS to estimate $\mathbf{W}$,
and use Lasso regression to estimate $\mathbf{P}$, independently.

NOTEARS:
 $\mathcal{L}(\mathbf{W}) = \frac{1}{2n}||\mathbf{X}-\mathbf{X}\mathbf{W}||^2_F + \lambda_{\mathbf{W}}||\mathbf{W}||_1. \text{s.t.,} \mathbf{W}\text{ is acyclic.}$

LASSO: $\mathcal{L}(\mathbf{P}) = \frac{1}{2n}||\mathbf{X}-\mathbf{A}\boxtimes\mathbf{M}\mathbf{P}||^2_F + \lambda_{\mathbf{P}}||\mathbf{P}||_1.$\\

\item DYNOTEARS~\cite{pamfil2020dynotears}: This is an extension of NOTEARS on time series. Compared with our method, it ignores the interactions between samples (i.e., $\mathbf{A}$).

Objective: $\mathcal{L}(\mathbf{W}, \mathbf{P}) = \frac{1}{2n}||\mathbf{X}-\mathbf{X}\mathbf{W}-\mathbf{M}\mathbf{P}||^2_F + \lambda_{\mathbf{W}}||\mathbf{W}||_1+ \lambda_{\mathbf{P}}||\mathbf{P}||_1. \text{s.t.,}\quad \mathbf{W}\text{ is acyclic.}$

\end{itemize}
For fully utilizing multiple-step graph series, these methods are also extended to tensor versions.
\begin{figure*}[htbp]
	\centering
	\includegraphics[width=16cm]{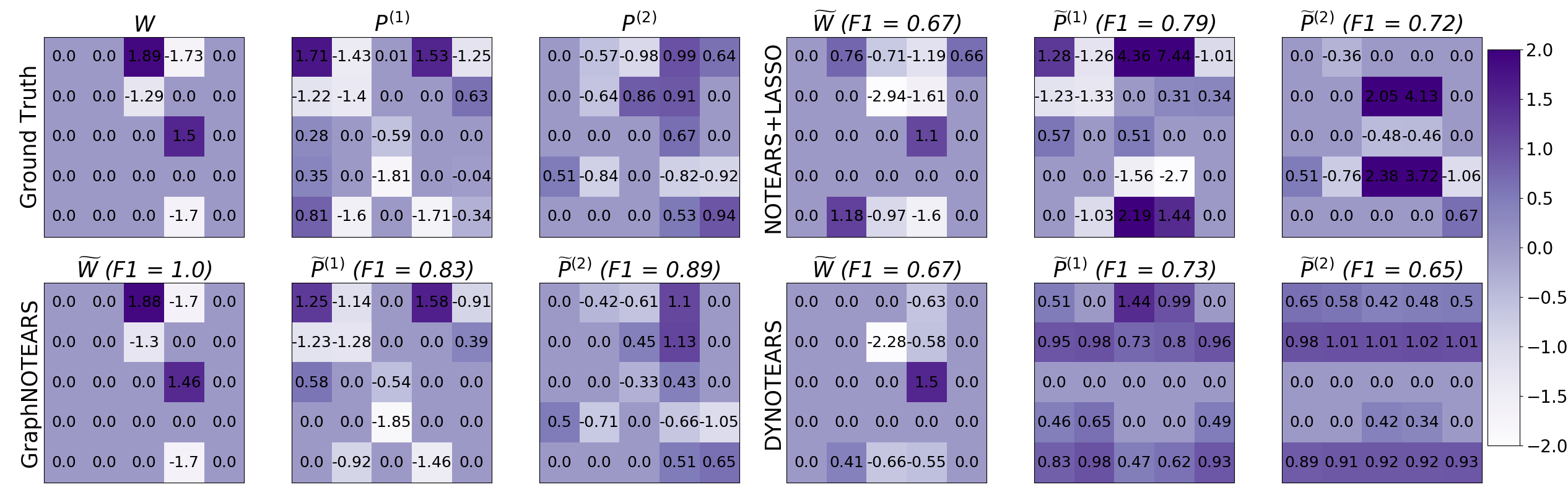}
	\caption{Example results for data with Gaussian noise, $n = 500$ samples, $d = 5$ variables at each timestamp, $T=7$ time-series, and $p = 2$ time-lagged graph effect order. Our algorithm recovers weights that are closer to the ground truth than baselines.}
	\label{fig:esitmated}
\end{figure*}

\textbf{Metrics.} We evaluate the learned binary intra-slice $\mathbf{\hat{W}}$ and inter-slice matrices $\mathbf{\hat{P}}$ separately by two common graph metrics: F1-score and Structural Hamming Distance (SHD)~\cite{zheng2018dags}.

\textbf{Experimental setup.} For all methods, we set hyperparameters $\lambda_{\mathbf{W}}=\lambda_{\mathbf{P}}=0.01$. For the weight thresholds, following~\cite{zheng2018dags}, we choose $\tau_{\mathbf{W}}=\tau_{\mathbf{P}}=0.3$ for all the methods. The
relative ranking of the three methods is not
sensitive to the weight thresholds according to our observation. For utilizing multiple-step graph series, we use the first $T-1$ timestamps to predict last $T-p$ timestamps, where $p$ time-lagged influence order is considered at each timestamp. For all experiments, we utilize 5 different random seeds to generate different datasets and initialize models, and report the mean value and 95\% confidence interval.
\begin{figure*}[htbp]
	\centering
	\includegraphics[ width=16cm]{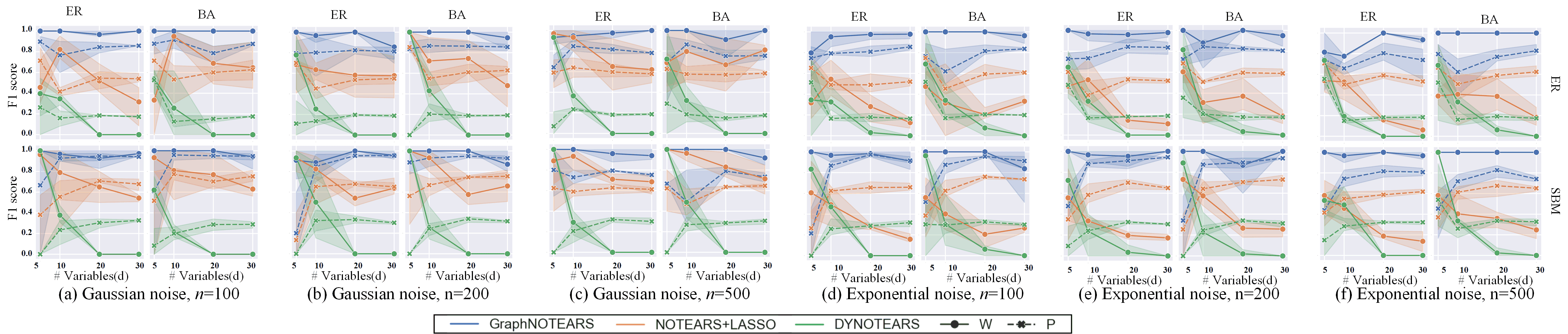}
	\caption{F1 scores (higher is better) for different noise models (Gaussian, Exponential) and different sample sizes ($n\in[100, 200, 500]$). The length of time-series is 7 and we consider 1-step time-lagged neighbor influence here. Each panel contains results for two different choices of intra-slice graphs (columns) and inter-slice graphs (rows). Every marker corresponds to the mean performance across 5 algorithm runs, where each on a different simulated dataset, and shade area means the  95\%  confidence interval. Continuous and dashed lines represent F1 scores for intra-slice and inter-slice edges, respectively.}
	\label{fig:f1 results}
\end{figure*}
\begin{figure*}[htbp]
	\centering
	\includegraphics[width=16cm]{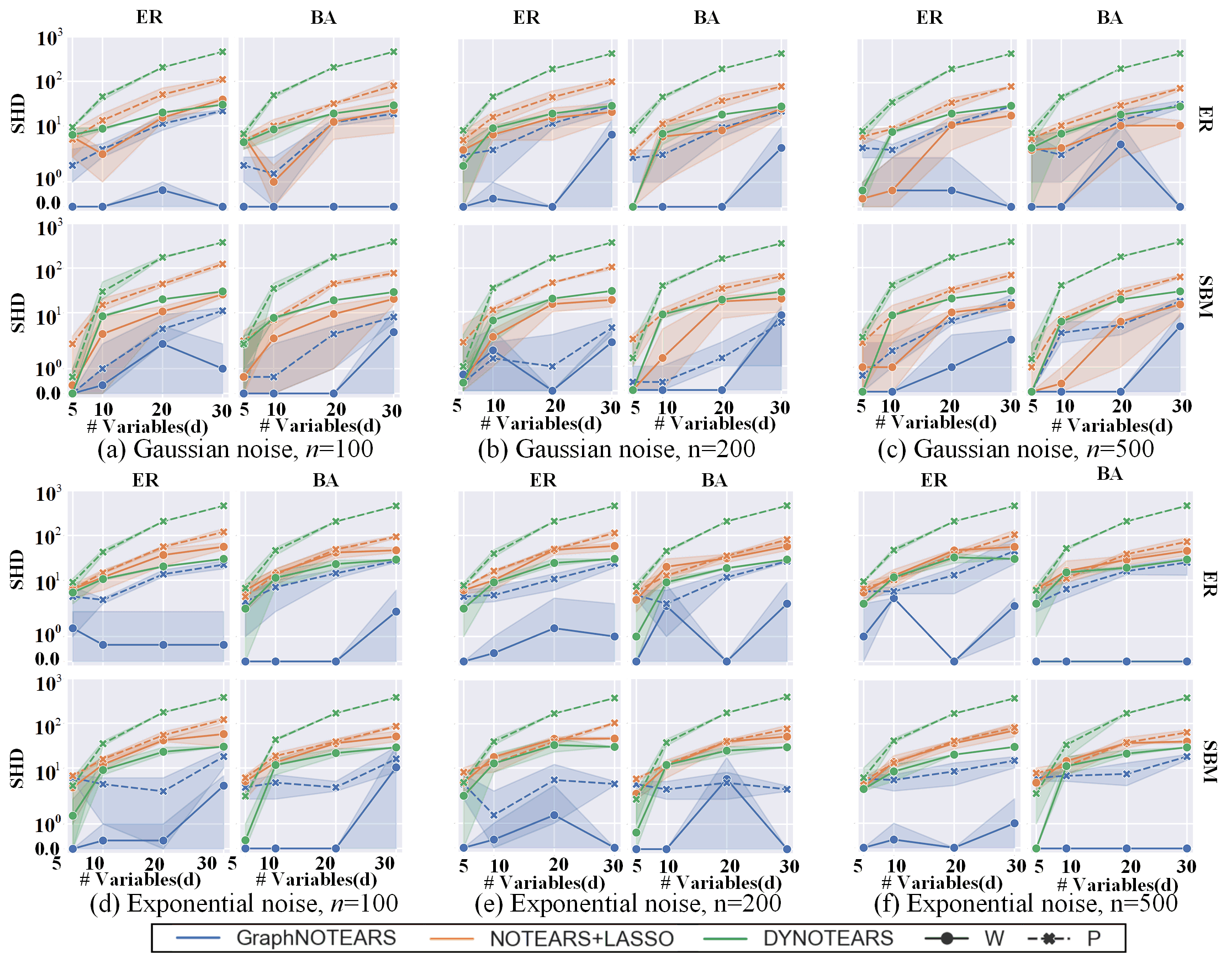}
	\caption{SHD scores. Illustrations are the same as Fig.~\ref{fig:f1 results}.}
	\label{fig:SHD results}
\end{figure*}
\begin{figure*}[htpb]
\centering
\subfigure[GraphNOTEARS on user dynamic graph.]{
\includegraphics[ width=7cm]{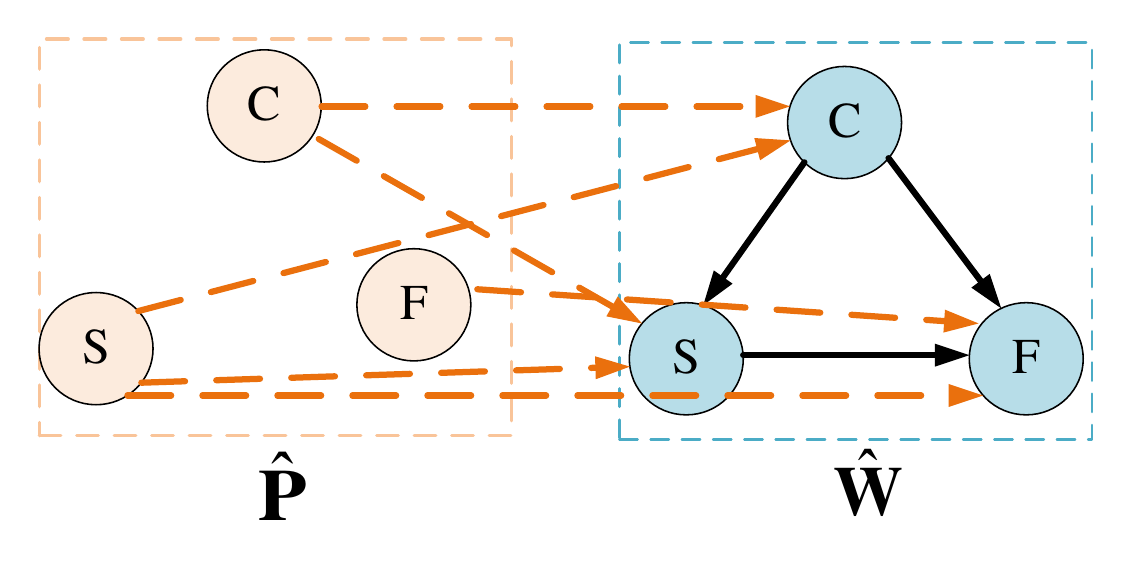}
\label{fig:gradient}
}
\subfigure[DYNOTEARS on user dynamic graph.]{
\includegraphics[width=7cm]{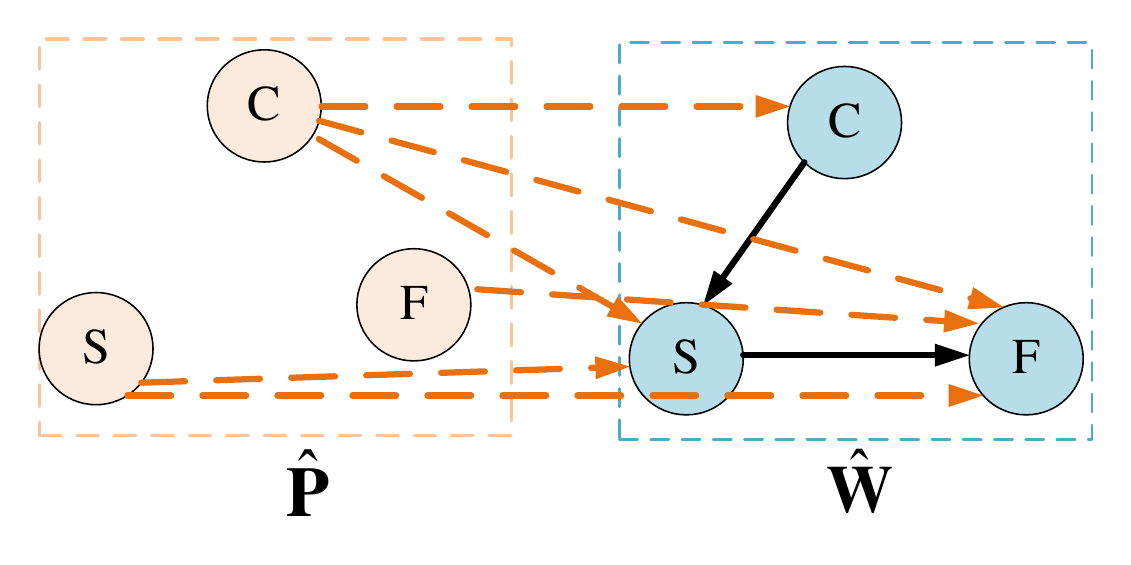}
\label{fig:gradient}
}

\subfigure[GraphNOTEARS on business dynamic graph.]{
\includegraphics[ width=7cm]{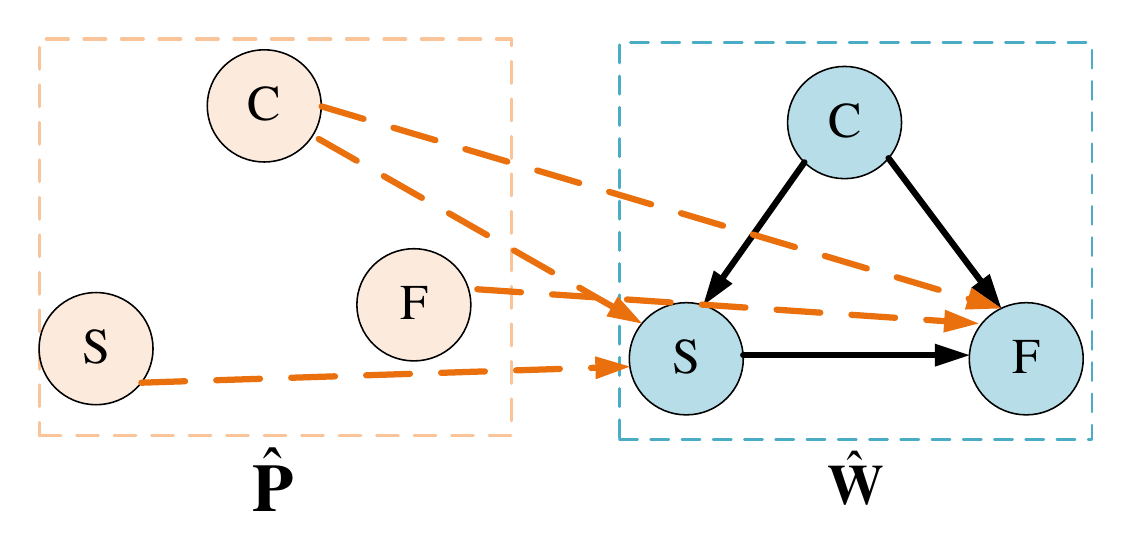}
\label{fig:gradient}
}
\subfigure[DYNOTEARS on business dynamic graph.]{
\includegraphics[width=7cm]{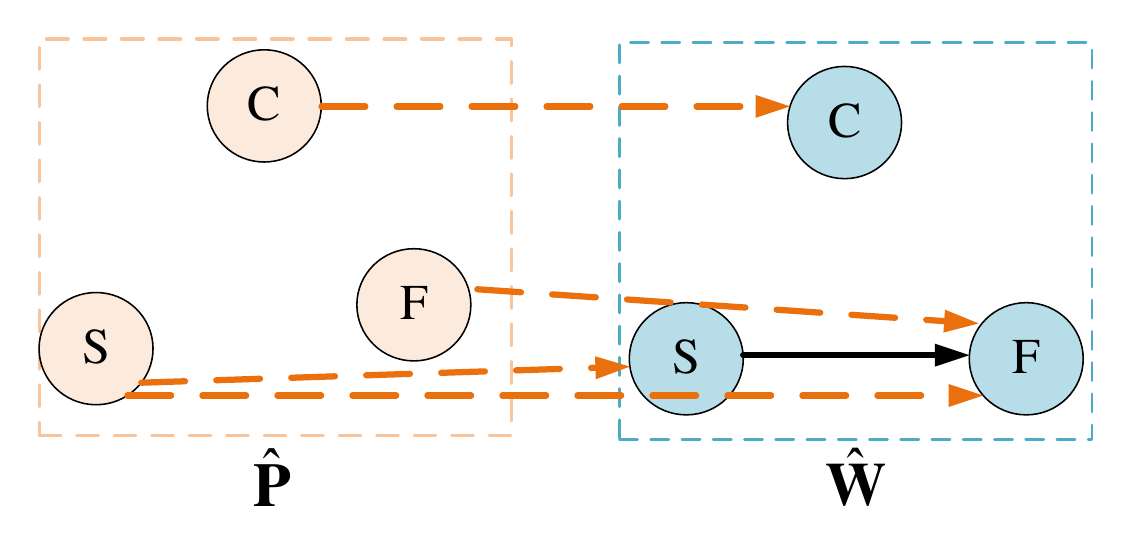}
\label{fig:gradient}
}
\caption{Estimated DAG on real-world Yelp dataset.}
\label{fig:realdata}
\end{figure*}
\subsection{Performance Evaluation}
We start by illustrating the estimated weighted matrices of GraphNOTEARS and baselines with the ground truth in Fig.~\ref{fig:esitmated}. The evaluation data is generated with Gaussian noise, $n = 500$ samples, $d = 5$ variables, $T=7$ time-series, and $p = 2$ time-lagged graph effect order. And the intra-slice and inter-slice are both generated with ER graph. As Fig.~\ref{fig:esitmated} shows, our estimated weights are much closer to the true weights
for both $\mathbf{W}$, $\mathbf{P}^{(1)}$ and $\mathbf{P}^{(2)}$ compared with baselines, where the intra-slice $\mathbf{W}$ is recovered perfectly (F1 score =1.0) and the inter-slice $\mathbf{P}^{(1)}$ and $\mathbf{P}^{(2)}$ are also better than baselines with a large margin. The results validate that our model is a general framework that could estimate the DAG of a dynamic graph with multiple timestamps and could simultaneously learn arbitrary time-lagged order influence. However, both NOTEARS+LASSO and DYNOTEARS could not achieve satisfying results. The reasons could be that NOTEARS+LASSO is a two-stage method that could not simultaneously consider the contemporaneous and time-lagged interaction influence to generate variables, while our model jointly models these two factors into a unified framework. Furthermore, DYNOTEARS only takes the time-lagged series information into consideration and ignores the interaction information. As we can see, if the generation of variables are determined by the neighborhood, it is necessary to 
incorporate the graph information into the model.

\par We present the F1-score and SHD results on the full setting in Fig.~\ref{fig:f1 results} and Fig.~\ref{fig:SHD results}, respectively. Note that, for simplicity, we set time-lagged graph order $p=1$ here. From the results, we have the following observations: (1) GraphNOTEARS is the best algorithm. In most cases, for inter-slice graph $\mathbf{W}$, our model nearly recovers the graph perfectly (F1 score$\approx$1.0 and SHD$\approx$0). For the intra-slice graph $\mathbf{P}$, our method also achieves satisfying results, outperforming baselines with a large margin. The phenomenon demonstrates that GraphNOTEARS could learn DAGs from dynamic graphs,  owing to the fact that it could comprehensively consider complex information. (2) Overall, with the number of variables increasing, especially in insufficient samples scenario, e.g., $n=100$, all the methods suffer from the degradation of performance. However, our model performs stable and still outperforms baselines with a large margin, indicating our model could handle the challenging high-dimensional scenario well. (3) No matter in what kind of underlying graphs or noise-type scenarios, our model could all achieve promising results. It indicates our model has the potential to deal with various scenarios which may encounter in real-world applications. (4) For SHD results, GraphNOTEARS requires less modification of edges to reach ground truth in all settings, further validating the effectiveness of the proposed method.

\section{Application on Real-world Datasets}
We consider applying GraphNOTEARS on two real-world dynamic graphs constructed from Yelp dataset~\cite{luca2016reviews}. \cite{anderson2012learning} introduced a toy SCM that embeds causal knowledge for
the Yelp example. That is, there are three random variables, i.e., the restaurant category $C$, Yelp star rating $S$, and customer flow $F$. There exist three directed edges that represent the three causal relationships between variables: (1) Restaurant category influences its Yelp rating. For example,
the average rating of fast-food restaurants is lower than that of high-end seafood restaurants. (2) Restaurant
category also influences its customer flow. For example, the average customer flow of high-end restaurants is lower than fast food. (3) Yelp rating of a restaurant influences its customer flow. According to this, we construct two dynamic graphs, i.e., the user graph and business graph, where its node features are these three variables. Particularly, we construct a user graph based on whether two users are friends on the Yelp platform. Then we take the time lag as one month and calculate the average category, the average Yelp star rating, and the average customer flow of the restaurants they have visited in this month as the node features. Here we consider 1-step time-lagged graph information. As the user's taste may be influenced by their friends, the generation of users' features should consider their friends' influence. For example, if a user's friend posts a positive review on this restaurant, the user will have a larger possibility to visit this restaurant. For the business graph, as the same category of restaurants may have the ``effect of agglomeration'' to influence each other, we add edges between the restaurants which have a similar category and close distance. Then we calculate the variables of the restaurants same as the user graph.

\par We apply GraphNOTEARS and DYNOTEARS on the constructed graphs and obtain the binary DAG via 0.1 threshold, shown in Fig.~\ref{fig:realdata}. For both two dynamic graphs, the estimated relationships of the intra-slice matrix $\hat{\mathbf{W}}$ discovered by GraphNOTEARS coincides with our prior knowledge, i.e., the three black directed edges among variables in Fig.~\ref{fig:realdata}(a)(c). However, DYNOTEARS could only discover partial edges, e.g., missing $C\rightarrow F$ in Fig.~\ref{fig:realdata}(b), and $C\rightarrow S$ and  $C\rightarrow F$ in Fig.~\ref{fig:realdata}(d). And we find that there are strong correlations between the same type of variables as illustrated in inter-slice  $\hat{\mathbf{P}}$ (e.g., the directed edge between the aggregated neighborhood category (orange S) with self category  (blue S)), which could be explained by the homophily influence of the graph. Overall, our model could discover an explainable DAG.

\section{Conclusion}
In this paper, we first study a new DAG learning diagram on dynamic graphs, which plays a vital role in understanding the node features generation mechanism. To handle such complex data, we propose a score-based DAG method to learn both intra-slice and inter-slice dependencies between variables simultaneously, considering both temporal and interaction information. The resulting method could deal with such a complex problem efficiently and has the potential for more complicated settings. Extensive experiments on both simulated and real-world datasets well demonstrate the effectiveness of the proposed method.
\section{Acknowledgments}
This work is supported in part by the National Natural Science Foundation of China (No. U20B2045, 62192784, 62172052, 62002029, 62172052, U1936014).
\bibliography{aaai23}

\end{document}